\documentclass[10pt,twocolumn,letterpaper]{article}
\usepackage[pagenumbers]{cvpr} 

\usepackage{graphicx}
\usepackage{amsmath}
\usepackage{amssymb}
\usepackage{booktabs}
\usepackage{comment}

\newsavebox\CBox
\def\textBF#1{\sbox\CBox{#1}\resizebox{\wd\CBox}{\ht\CBox}{\textbf{#1}}}
\newcommand{\bestscore}[1]{\textcolor{black}{\textBF{#1}}}

\usepackage[pagebackref,breaklinks,colorlinks]{hyperref}

\usepackage[capitalize]{cleveref}
\crefname{section}{Sec.}{Secs.}
\Crefname{section}{Section}{Sections}
\Crefname{table}{Table}{Tables}
\crefname{table}{Tab.}{Tabs.}

\begin{document}

\title{ClipCrop: Conditioned Cropping Driven by Vision-Language Model}

\author{Zhihang Zhong$^{1,2}$\and
Mingxi Cheng$^{3}$\and
Zhirong Wu$^{3}$\and
Yuhui Yuan$^{3}$\and
Yinqiang Zheng$^{1}$\and
Ji Li$^{3}$\and
Han Hu$^{3}$\and
Stephen Lin$^{3}$\and
Yoichi Sato$^{1}$\and
Imari Sato$^{1,2}$\and
$^1$The University of Tokyo, Japan\qquad
$^2$National Institute of Informatics, Japan\qquad
$^3$Microsoft Research
\\
{\tt\small \{zhong@is.s,yqzheng@ai,ysato@iis\}.u-tokyo.ac.jp\qquad imarik@nii.ac.jp}\\ {\tt\small \{mingxicheng,wuzhiron,yuhui.yuan,jili5,hanhu,stevelin\}@microsoft.com}
}

\maketitle

\begin{abstract}
   Image cropping has progressed tremendously under the data-driven paradigm. However, current approaches do not account for the intentions of the user, which is an issue especially when the composition of the input image is complex. Moreover, labeling of cropping data is costly and hence the amount of data is limited, leading to poor generalization performance of current algorithms in the wild. In this work, we take advantage of vision-language models as a foundation for creating robust and user-intentional cropping algorithms. By adapting a transformer decoder with a pre-trained CLIP-based detection model, OWL-ViT, we develop a method to perform cropping with a text or image query that reflects the user's intention as guidance. In addition, our pipeline design allows the model to learn text-conditioned aesthetic cropping with a small cropping dataset, while inheriting the open-vocabulary ability acquired from millions of text-image pairs. We validate our model through extensive experiments on existing datasets as well as a new cropping test set we compiled that is characterized by content ambiguity.
\end{abstract}

\section{Introduction}
\label{sec:intro}
\begin{figure*}[!th]
  \centering
  \includegraphics[width=\linewidth]{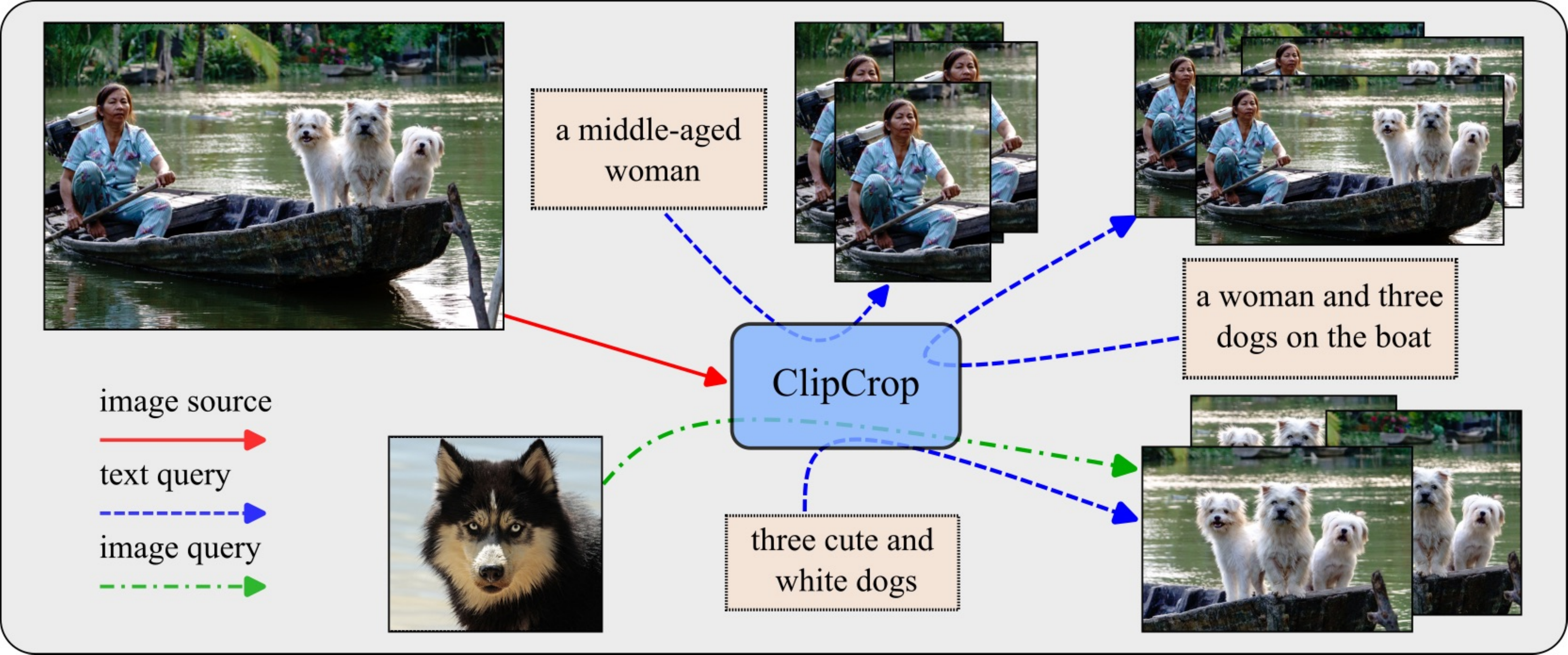}
  \caption{\textbf{The concept of conditioned smart cropping.} Given an image with complex composition, the proposed ClipCrop can accept different text queries, such as ``a middle-aged woman'', ``three cute and white dogs'' and ``a woman and three dogs on the boat'', as conditions to return corresponding aesthetic-aware crops. In addition, ClipCrop can also accept an image as a query condition for cropping.}
  \label{fig:teaser}
\end{figure*}

Images often need to be cropped when circulating among different devices and social media apps. The key requirement is to maintain the aesthetics of the image, while not ruining the intended content. Many datasets~\cite{yan2013learning,fang2014automatic,chen2017quantitative,wei2018good,zeng2020grid} with aesthetics-aware annotations have been constructed for benchmarking automatic cropping algorithms. Nevertheless, how to accurately reflect users' intentions of image cropping has not been well studied.

In recent years, data-driven learning has dramatically boosted the performance of image cropping methods. Existing methods can be broadly divided into anchor (sliding window) evaluation methods~\cite{nishiyama2009sensation,chen2017learning,wei2018good,zeng2020grid,li2020composing,pan2021transview} and coordinate regression methods~\cite{li2020learning,lu2020weakly,hong2021composing,jia2022rethinking}. There are two problems with the practical application of these methods. First, desirable crop of a given image highly depends on a user's intention, such as how the user want to use the resulting image or from what perspective the user sees the image. If the composition of the input image is complex, such as the presence of many objects, there is no way for general cropping methods to accurately predict the user's intention, but only to make random guesses. Second, due to the tedious process of labeling a cropping dataset, the size of existing datasets is rather small. This results in current approaches being poorly generalized and often failing when faced with unseen scenarios.

The success of vision-language modeling (VLM)~\cite{radford2021learning,minderer2022simple} gives us inspiration to address the above two challenges. Today, we have access to considerable amounts of multimodal data, \eg, images with labels or titles. To tackle the first challenge, text can be introduced as additional input to reflect the user's intention, which naturally eliminates content ambiguity during learning. As for the second challenge, transferring VLM's open-vocabulary knowledge from millions of image-text pairs to cropping algorithms is a promising but unexplored direction for more effective cropping in the wild.

In this work, we propose ClipCrop: a vision-language model for generating aesthetic image crops based on text or image queries.
The proposed model is built on OWL-ViT~\cite{minderer2022simple}, which is one of the state-of-the-art open-vocabulary object detection models. We adopt a transformer decoder like DETR~\cite{carion2020end} to learn aesthetic-aware offsets given the initially predicted bounding box from OWL-ViT. 
Since no text description is available for cropping training datasets,
we provide annotations by complementing GAIC~\cite{zeng2020grid} with text descriptions for each image. 
When fine-tuning our model, we adopt a mosaics composition training strategy, which encourages the model to localize from text queries. 
The improved model can comprehensively understand input text while retaining the ability of one-shot image-conditioned querying. Taking the source image in Fig.~\ref{fig:teaser} as an example, there are multiple contents of interest in the image, \eg, a woman, three dogs, or a woman and three dogs on the boat. The proposed ClipCrop takes the user text/image query as input to understand the user's intent and generates several crop solutions.

Horanyi-PR~\cite{horanyi2022repurposing} is the only cropping evaluation benchmark that provides text descriptions, with a single sentence per image.
To emphasize the content ambiguity problem for cropping, we further provide an evaluation benchmark, TextCrop for short, with two distinct text annotations and corresponding aesthetics-aware bounding box annotations for each image. This new evaluation benchmark consists of about 200 crowded photos containing multiple contents. Experimental results demonstrate that the proposed ClipCrop achieves significant improvement in terms of average and maximum IoUs in comparison to the prior arts on the Horanyi-PR and the proposed TextCrop test sets. Even in the general cropping setting of GAIC~\cite{zeng2020grid}, we show that our baseline model performs at a SoTA level and reveal the mismatch between evaluation on this dataset and practical applications.

Our contributions can be summarized as follows.
\begin{itemize}
    \item We are the \textit{first} to adapt VLMs for conditioned image cropping. The proposed method makes it possible to let the user provide a text or image query to crop, better reflecting the user's intention.
    \item We present a pipeline to generalize smart cropping models that are trained with expensive yet limited human-annotated data to unseen photos in the wild.
    \item We verify the performance of our ClipCrop via extensive experiments, and provide a novel evaluation dataset focusing on the challenge of content ambiguity.
\end{itemize}


\section{Related Works}
\label{sec:related}

In this section, we review the related literature on image cropping, visual grounding, and vision-language modeling.

\subsection{Image Cropping}
\label{sec:image_cropping}
Image cropping algorithms aim to improve the aesthetic quality of images by removing redundant content and can serve many image editing tasks, \eg, thumbnail generation~\cite{huang2015automatic}. The algorithms for cropping can be roughly divided into anchor evaluation methods and coordinate (bounding box) regression methods. 

Anchor evaluation is a two-step process of generating and scoring a crop proposal. The first step can be realized through sliding windows~\cite{kang2014convolutional,lu2014rapid,kong2016photo,chen2017learning}, predefined anchor boxes~\cite{wei2018good} similar to detection models~\cite{liu2016ssd}, or grid anchors~\cite{zeng2020grid}. For the second step, distinct quality scorers~\cite{nishiyama2009sensation} are created to rank the crop proposals. Many works like Ni~\etal~\cite{ni2013learning} propose algorithms to mine the implicit composition rules behind photos with high aesthetic quality, such as the ``rule of thirds'', ``rule of space'', and ``rule of symmetry'' popular in the field of photography. Fang~\etal~\cite{fang2014automatic} build a system consisting of three models that learn a visual composition rule, boundary simplicity rule, and content preservation rule for comprehensive scoring. Also, saliency maps~\cite{fang2014automatic} and foreground identity~\cite{huang2015automatic} are often considered as critical factors for scoring. Yan~\etal\cite{yan2013learning} highlight the importance of removed or changed content from the original image for crop quality assessment, while Zeng~\etal~\cite{zeng2020grid} adopt an RoD operation to consider the region of discard. Chen~\etal~\cite{chen2017learning} presume that a professional photograph on a website is generally better than its corresponding crop in terms of aesthetics. Therefore, instead of direct scoring, they propose VFN to learn how to compare the quality between two input images by harvesting many examples of unambiguous paired aesthetic rankings from a website. Li~\etal~\cite{li2020composing} exploit mutual relations between good crops and bad crops with a graph convolution operation. More recently, TransView~\cite{pan2021transview} provides a transformer-based solution that achieves strong performance and presents interesting visualizations about the attraction and repulsion between visual elements to explain the reasons behind the scores.

For coordinate regression methods, the goal is to directly predict the cropping boxes. Guo~\etal~\cite{guo2018automatic} propose a cascaded pipeline to iteratively regress the final box, while Li~\etal~\cite{li2018a2} introduce a reinforcement learning framework to mimic human adjustment decisions for dynamic adjustment of the predicted box. In addition, Li~\etal propose Mars~\cite{li2020learning} to realize arbitrary aspect ratio cropping box prediction by meta learning. Instead of fully supervised learning, Lu~\etal~\cite{lu2020weakly} propose a weakly supervised cropping framework guided by the distribution dissimilarity between the predicted cropped images and high-quality images, which eliminates the need for ground-truth boxes annotated by humans. Then, Hong~\etal propose CACNet~\cite{hong2021composing} that combines an anchor-point-based cropping branch and an interpretable composition branch to learn the offsets of each anchor point to the ground-truth boxes. Inspired by recent advances in object detection~\cite{carion2020end}, Jia~\etal~\cite{jia2022rethinking} explicitly address the diversity problem in coordinate regression-based cropping methods by learning in a set prediction fashion.

The closest work to ours is CAGIC~\cite{horanyi2022repurposing}, which to the best of our knowledge is the first text-conditioned image cropping model. Horanyi~\etal~\cite{horanyi2022repurposing} propose to directly re-purpose pre-trained networks to realize text-conditioned cropping without any fine-tuning. We begin with a VLM perspective and implement more general conditioned cropping, including text querying and image querying. More importantly, our model inherits an open vocabulary capability, which is more stable and accurate for both the dataset presented with GAIC and in practical applications.

\subsection{Visual Grounding}
\label{sec:visual_grounding}

The basic goal of visual grounding~\cite{karpathy2014deep} is to predict and identify a set of bounding boxes surrounding the target objects that are referred to by the given text query. According to the task formulations, we can categorize the existing efforts into two paths, namely phrase grounding~\cite{plummer2015flickr30k} which maps a sentence consisting of multiple entities into a set of bounding boxes, and referring expression comprehension~\cite{yu2016modeling,mao2016generation} which only maps a referring expression to a single bounding box. Most of the previous works~\cite{yu2018mattnet,lu2019vilbert,su2019vl,lu202012,li2020unicoder,chen2020uniter,yang2020improving,deng2021transvg,shi2022dynamic} address the visual grounding tasks by exploiting a well-trained existing object detector (\eg, DarkNet~\cite{redmon2018yolov3} and DETR~\cite{carion2020end}) to generate the candidate bounding boxes and then search for the target based on semantic similarities between region features and textual features. Recently, MDETR~\cite{kamath2021mdetr}, GLIP~\cite{li2022grounded,zhang2022glipv2}, and OWL-ViT~\cite{minderer2022simple} choose to embrace a scheme that performs end-to-end training and alignment between texts and objects in the image directly. The key difference between visual grounding and conditioned image cropping is that the latter requires consideration of aesthetic elements, and often the box may include multiple objects combined with a background. In this work, we construct our approach based on the very recent OWL-ViT~\cite{minderer2022simple} that already exploits rich datasets for visual grounding including OpenImages-V$4$~\cite{kuznetsova2020open} and Visual Genome~\cite{krishna2017visual}.

\subsection{Vision-Language Modeling}
\label{sec:related_vlm}

Vision-language modeling (VLM) attempts to exploit one data modality as the supervision for the other modality. The research on VLM follows the success of BERT~\cite{devlin2018bert}, where image-conditioned BERT models~\cite{lu2019vilbert,su2019vl} are developed to facilitate the downstream task of visual question answering and image captioning. Later works~\cite{li2020oscar,wang2021simvlm,li2021align,kim2021vilt,li2020unicoder,chen2020uniter} further incorporate image masked modeling, image text matching, and multimodal fusion for visual grounding and image and text retrieval. BERT-like models require fine-tuning in order to reach satisfactory results.

CLIP~\cite{radford2021learning} revolutionized the field by enabling zero-shot capabilities through large-scale image-caption contrastive learning. The pre-trained CLIP model shows a strong generalization ability to open-set problems with prompt tuning~\cite{lester2021power} and prompt engineering~\cite{liu2021pre}. The work following CLIP extends the zero-shot capability for detection~\cite{minderer2022simple,zhong2022regionclip}, segmentation~\cite{xu2022groupvit,ghiasi2021open,rao2022denseclip}, and many other applications~\cite{wu2022wav2clip,xu2021videoclip,sanghi2022clip}. Models that improve upon CLIP include LiT~\cite{zhai2022lit}, Align~\cite{jia2021scaling}, UniCL~\cite{yang2022unified}, and BASIC~\cite{pham2021combined}. Our work leverages the language interface of CLIP for smart cropping, enabling the user to express intention with regard to image regions.


\section{Methodology}
\label{sec:method}
In this section, we first review the working mechanism of OWL-ViT~\cite{minderer2022simple}, which serves as an encoder for our model. Then, we present the structure of our ClipCrop model. Finally, we further elaborate on the training strategy with conditioned cropping.

\begin{figure*}[!t]
  \centering
  \includegraphics[width=\linewidth]{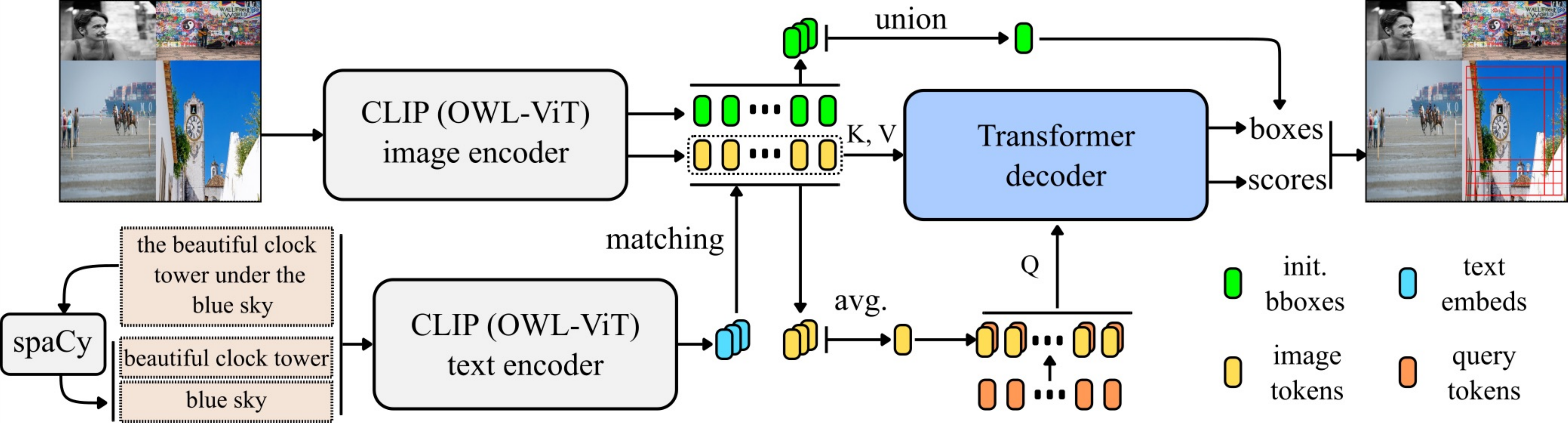}
  \caption{\textbf{The architecture of ClipCrop.} In addition to the full input text, ClipCrop uses spaCy~\cite{Honnibal_spaCy_Industrial-strength_Natural_2020} to extract keywords and then generates a series of text embeddings via the text encoder of OWL-ViT~\cite{minderer2022simple}. These text embeddings are used to match image tokens and corresponding initial bounding boxes from the image encoder of OWL-ViT. To achieve conditioned cropping, a transformer decoder similar to DETR~\cite{carion2020end} is introduced to regress a series of offsets for the union of the selected initial boxes and predict the corresponding quality scores.}
  \label{fig:ClipCrop}
\end{figure*}

\subsection{Review of OWL-ViT}
\label{sec:review}

OWL-ViT uses a standard Vision Transformer as the image encoder $\mathcal{F}_{img}$ and the text encoder $\mathcal{F}_{txt}$. Training is conducted in two stages. In the first stage, $\mathcal{F}_{img}$ and $\mathcal{F}_{txt}$ are contrastively pre-trained with large-scale image-text pairs following the standard CLIP~\cite{radford2021learning} process. In the second stage, the pooling operation and the final projection layer are removed from $\mathcal{F}_{img}$. Then, the output image tokens $\{t^{n}_{img}\}$ are projected to per-object classification embeddings $\{e^{n}_{cls}\}$ by a linear layer as well as to the corresponding bounding boxes $\{b^{n}\}$ by a MLP.
In the case of using ViT-B/32 backbone, $n \in \{1,\cdots,576\}$.
This stage is trained using publicly available detection datasets including OpenImages-V$4$~\cite{kuznetsova2020open}, Objects 365 (O365)~\cite{shao2019objects365}, and Visual Genome~\cite{krishna2017visual}. In inference, a text embedding $e_{txt}$ from $\mathcal{F}_{txt}$ is used as guidance to match the most similar classification embedding. Then, the corresponding box of the matched classification embedding becomes the predicted box.

\subsection{Structure of ClipCrop}
\label{sec:structure}
ClipCrop processes the image tokens $\{t^{n}_{img}\}$ and the predicted boxes $\{b^{n}_{ini}\}$ (renaming of $\{b^{n}\}$) from the image encoder of OWL-ViT as the initial ingredients, as illustrated in Fig.~\ref{fig:ClipCrop}. The target of OWL-ViT is visual grounding. This means that the predicted boxes are for single objects without any consideration of aesthetics. To adapt OWL-ViT to conditioned cropping, in addition to the original text input, ClipCrop further uses spaCy~\cite{Honnibal_spaCy_Industrial-strength_Natural_2020} to extract keywords from it and generate several text embeddings $\left\{e^{w}_{txt}\,|\,w\in\{1,\cdots,W\}\right\}$ via the text encoder. Then ClipCrop follows the matching mechanism of OWL-ViT by using $\{e^{w}_{txt}\}$ to select a series of image tokens $\{t^{w}_{img}\}$ and the corresponding bounding boxes $\{b^{w}_{ini}\}$. ClipCrop introduces a new DETR-like transformer decoder $\mathcal{F}_{dec}$ to regress a set of offsets $\{b^{m}_{ofs}\}$ for the union box $b_{u}$ of $\{b^{w}_{ini}\}$ and the corresponding quality scores $\{s^{m}\}$, as follows:
\begin{equation}
    \{b^{m}_{ofs}\}, \{s^{m}\}  = \mathcal{F}_{dec}\left(Q,K,V\right),
\end{equation}
where $m\in\{1,\cdots,M\}$ and $M=90$ by default, as in Jia~\etal~\cite{jia2022rethinking}. Key $K$ and value $V$ are from complete image tokens $\{t^{n}_{img}\}$, and the query $Q$ is the sum of the average of selected image tokens $\{t^{w}_{img}\}$ and $M$ learnable query tokens $\{t^{m}_{qry}\}$:
\begin{equation}
    Q = \left\{ t^{m}_{qry} + \frac{1}{W}\sum\nolimits^{W}_{w=1}t^{w}_{img} \right\}.
\end{equation}
The final predicted boxes are the sum of the union box and the regressed offsets:
\begin{equation}
    \{b^{m}_{pred}\} = b_{u} + \left\{b^{m}_{ofs}\right\}.
\end{equation}

\subsection{Training Strategy}
\label{sec:training}

Our training strategy basically follows that of Jia~\cite{jia2022rethinking} to supervise the predicted boxes and scores. The Hungarian algorithm~\cite{kuhn1955hungarian} is used to find a bipartite matching between predicted bounding boxes and high-quality ground-truth boxes. The high-quality boxes in the GAIC~\cite{zeng2020grid} dataset represent boxes with a score of no less than 4. The scores in this dataset range from 1 to 5, with higher scores indicating better. Regarding quality guidance, we also leverage local redundancy~\cite{zeng2019reliable}, \ie, human perception is insensitive to small changes of the crop, to achieve label smoothing. Specifically, if there exists a ground-truth box that has high IoU (IoU $\geq 0.9$) with a predicted box, the quality score of the predicted box can be supervised using the quality score of that ground-truth box.

To train a conditioned cropping algorithm, there are still several problems. First, there are no paired texts for training. Therefore, we manually annotated a text for each image of GAIC~\cite{zeng2020grid} using a crowdsourcing platform. With this enhanced GAIC dataset, there is sufficient data for the community to train conditioned cropping algorithms. Second, because the ground-truth boxes of the GAIC~\cite{zeng2020grid} dataset are generated with a fixed grid pattern, the boxes generated by direct training using a single image will have bias. Hence, we deploy mosaics augmentation~\cite{bochkovskiy2020yolov4} to overcome this issue. We randomly sample single images, $2\times2$ grids, and $3\times3$ grids with equal probability, while the input text and ground-truth only belong to a single image. However, mosaic augmentation can introduce ambiguity for the input text when different images in the composed mosaic image have similar content. To deal with this, we filter out the initial boxes $\{b^{w}_{ini}\}$ that are far from the box with the highest quality score in the target image or contain less than half the area of that box. Regarding to more training details, please refer to the supplementary materials.

\section{Experiments}
\label{sec:experiments}

\subsection{Dataset}
\label{sec:dataset}

Many benchmark datasets~\cite{yan2013learning,fang2014automatic,chen2017quantitative,wei2018good,zeng2020grid} have been constructed for image cropping. Among these datasets, CPC~\cite{wei2018good} and GAIC~\cite{zeng2020grid} are densely annotated datasets, while only GAIC is dense enough to directly train a transformer decoder for predicting a set of boxes, as indicated by Jia~\etal~\cite{jia2022rethinking}. We train ClipCrop on GAIC~\cite{zeng2020grid} and evaluate it on GAIC's test set, Horanyi-PR~\cite{horanyi2022repurposing}, and the proposed TextCrop. 

\paragraph{GAIC~\cite{zeng2020grid}} This dataset contains 2636, 200, and 500 images for training, validation, and testing, respectively. Each image has about 90 crop proposals generated along a predefined grid pattern, which is intended to reflect the non-uniqueness of image cropping in practice. For each crop proposal, there is also a quality score ranging from 1 to 5. To train the text-conditioned cropping algorithm, we complemented GAIC by labeling each image with a text description.

\paragraph{Horanyi-PR~\cite{horanyi2022repurposing}} This dataset is designed to evaluate the ability to automatically crop images for a given caption. There are 100 images in total for testing, without training and evaluation sets. Each image has a text description and eight aesthetics-aware bounding boxes annotated by different human annotators.

\paragraph{TextCrop} We propose a new cropping test set that is characterized by content ambiguity, named TextCrop for short. TextCrop has 208 crowded photos with various scenarios and styles. We labeled each image with two different text descriptions and four aesthetics-aware bounding boxes from different human annotators for each text. For more details, please see our supplementary materials.

\begin{table}[t]
\caption{\textbf{Comparison on the Horanyi-PR~\cite{horanyi2022repurposing} dataset.}}
	\label{tab:horanyi-pr}
	\centering
	\setlength{\tabcolsep}{8pt}
	\begin{tabular}{lcc}
		\toprule
		Method & IoU-Mean & IoU-Max \\
		\midrule
		\textit{Visual Grounding} & & \\
		ReSC-Large~\cite{yang2020improving} & 0.4531 & 0.6079 \\ 
		TransVG~\cite{deng2021transvg} & 0.4398 & 0.6078 \\ %
		OWL-ViT~\cite{minderer2022simple} & 0.4131 & 0.5789 \\ %
		\midrule
		\textit{Cropping} & & \\
		A2-RL~\cite{li2018a2} & 0.3017 & 0.4506 \\ %
		VPN~\cite{wei2018good} & 0.3231 & 0.4830 \\ %
		VEN~\cite{wei2018good} & 0.3394 & 0.4882 \\ %
		Li~\etal~\cite{li2020composing} & 0.3357 & 0.4805 \\ %
		GAIC~\cite{zeng2020grid} & 0.3407 & 0.4863 \\ %
		CACNet~\cite{hong2021composing} & 0.3385 & 0.4899 \\ %
		Jia~\etal~\cite{jia2022rethinking} & 0.3333 & 0.4818 \\ %
		\midrule
		\textit{Conditioned Cropping} & & \\
		CAGIC~\cite{horanyi2022repurposing} & 0.4160 & - \\ 
		GAIC~\cite{zeng2020grid}+TransVG~\cite{deng2021transvg} & 0.4914 & 0.6290 \\
		ClipCrop & \bestscore{0.5551} & \bestscore{0.7284} \\
		\bottomrule
	\end{tabular}
\end{table}

\begin{figure*}[!t]
  \centering
    \includegraphics[width=\linewidth]{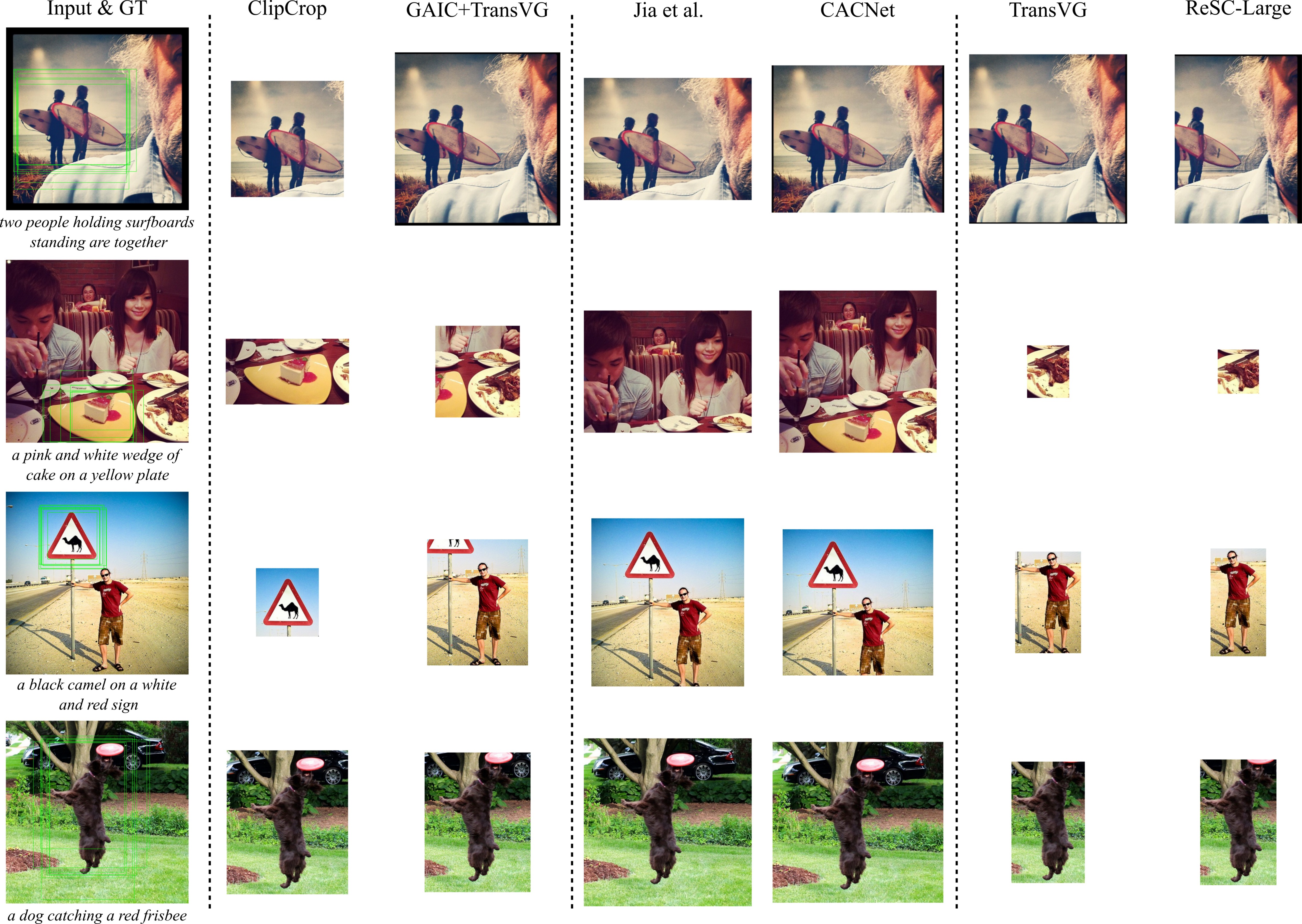}
  \caption{\textbf{Visual results on Horanyi-PR~\cite{horanyi2022repurposing}.} The first column presents the image and text description of the input. The green bounding boxes in the image represent the ground-truth (GT) crops. We show the results of our ClipCrop, cascaded scheme (GAIC~\cite{zeng2020grid}+TransVG~\cite{deng2021transvg}), general cropping algorithms (Jia~\etal~\cite{jia2022rethinking}, CACNet~\cite{hong2021composing}), and visual grounding algorithms (TransVG~\cite{deng2021transvg} and ReSC-Large~\cite{yang2020improving}) sequentially in the subsequent columns. ClipCrop outperforms other methods by accurately reflecting the intent of the text and ensuring the aesthetic quality of the crop.}
  \label{fig:horanyi-pr_results}
\end{figure*}

\subsection{Results on Horanyi-PR}
\label{sec:results_horanyi-pr}

First, we compare ClipCrop to the prior arts for visual grounding, general cropping, and text-conditioned cropping on the Horanyi-PR dataset~\cite{horanyi2022repurposing}. The visual grounding algorithms that are included, such as ReSC-Large~\cite{yang2020improving} and TransVG~\cite{deng2021transvg}, mainly deal with expression comprehension and return a single bounding box. Regarding the original OWL-ViT~\cite{minderer2022simple}, we take the bounding box with the highest matching score as the output. The compared general cropping algorithms cover both anchor evaluation methods~\cite{wei2018good,li2020composing,zeng2020grid} and coordinate regression methods~\cite{li2018a2,hong2021composing,jia2022rethinking}. All these algorithms are trained in their respective official settings and then tested on the unseen Horanyi-PR dataset. To the best of our knowledge, the only text-conditioned cropping algorithm that exists is the baseline algorithm CAGIC~\cite{horanyi2022repurposing} provided with this dataset. Since the implementation code of CAGIC is not published, we directly report the performance given in their paper. In addition, we implemented a cascaded scheme by connecting GAIC~\cite{zeng2020grid} and TransVG~\cite{deng2021transvg}. The returned crop is the one with the highest GAIC score and can cover the crop generated by TransVG. In this experiment, ClipCrop, CAGIC, the cascaded scheme (GAIC+TransVG), and visual grounding algorithms can utilize the text input of this test set, while the general cropping algorithms cannot.

\begin{figure*}[!t]
  \centering
    \includegraphics[width=\linewidth]{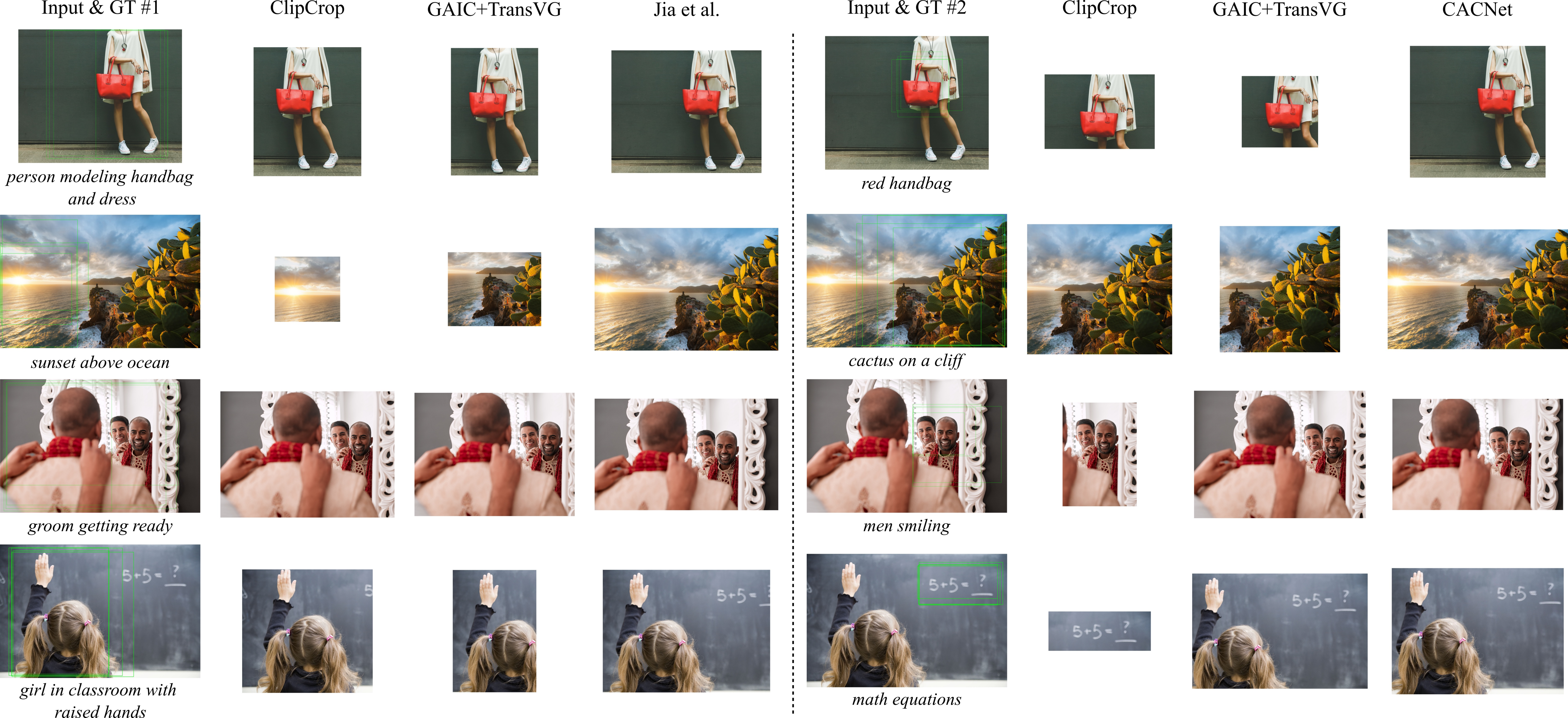}
  \caption{\textbf{Visual results on TextCrop.} The left side and the right side show the results of the same image with different text input as query. The green bounding boxes in the input represent the ground-truth. We show the results of our ClipCrop, cascaded scheme (GAIC~\cite{zeng2020grid}+TransVG~\cite{deng2021transvg}), and general cropping algorithm (Jia~\etal~\cite{jia2022rethinking}) for reference. ClipCrop can robustly discern intentions from different texts and crop accordingly.}
  \label{fig:textcrop_results}
\end{figure*}

The quantitative results are shown in Table~\ref{tab:horanyi-pr}. In addition to the average IoU (IoU-Mean) in~\cite{horanyi2022repurposing}, we calculate the maximum IoU (IoU-Max) between the bounding boxes labeled by human annotators and the predicted bounding box to reflect the performance. We can see that ClipCrop outperforms CAGIC, general cropping algorithms and visual grounding algorithms by a large margin.
We find that the visual grounding algorithms are superior to general cropping algorithms and even CAGIC. The reason why visual grounding algorithms are better than the general cropping algorithms is that the text descriptions of this dataset are biased towards small objects in the image, while general cropping algorithms tend to return a large crop that considers the global image. The reason that even CAGIC is inferior to visual grounding algorithms is due to its algorithm design.
CAGIC combines a pre-trained aesthetic evaluation network~\cite{chen2017learning} and a pre-trained image captioning network~\cite{xu2015show} to optimize a bilinear sampler to find a suitable crop, which leads to a low sum of aesthetic loss and caption loss. The match between the input text and the caption generated by~\cite{xu2015show} is not accurate, resulting in the generated bounding box having a rather poor performance compared to the tight bounding box generated by the visual grounding algorithms, which have a more accurate understanding of the input text.
The cascade scheme can combine the advantages of visual grounding algorithms and anchor evaluation cropping algorithms, but the improvement is limited.

We show qualitative results in Fig.~\ref{fig:horanyi-pr_results}. The case in the first row demonstrates that ClipCrop can effectively reflect the user's intention by removing extraneous background. The examples in the second and third rows demonstrate the powerful generalization capabilities of ClipCrop. ClipCrop can understand complicated descriptions such as ``a pink and white wedge of cake on a yellow plate'' and ``a black camel on a white and red sign'', while TransVG and ReSC-Large are unable to localize the expected area. In the case that the main content of the image and input text are more consistent, such as the fourth row, the results of ClipCrop and clipping algorithms are closer.

\begin{table}[t]
\caption{\textbf{Comparison on our TextCrop dataset.}}
\label{tab:textcrop}
	\centering
	\setlength{\tabcolsep}{8pt}
	\begin{tabular}{lcc}
		\toprule
		Method & IoU-Mean & IoU-Max \\
		\midrule
		\textit{Visual Grounding} & & \\
		ReSC-Large~\cite{yang2020improving} & 0.3116 & 0.3784 \\ 
		TransVG~\cite{deng2021transvg} & 0.3363 & 0.4056 \\ %
		OWL-ViT~\cite{minderer2022simple} & 0.3923 & 0.4758 \\ %
		\midrule
		\textit{Cropping} & & \\
		A2-RL~\cite{li2018a2} & 0.3919 & 0.4800 \\ %
		VPN~\cite{wei2018good} & 0.4052 & 0.4934 \\ %
		VEN~\cite{wei2018good} & 0.3758 & 0.4531 \\ %
		Li~\etal~\cite{li2020composing} & 0.4185 & 0.5034 \\ %
 		GAIC~\cite{zeng2020grid} & 0.4141 & 0.4978 \\ %
		CACNet~\cite{hong2021composing} & 0.4271 & 0.5162 \\ %
		Jia~\etal~\cite{jia2022rethinking} & 0.4317 & 0.5219 \\ %
		\midrule
		\textit{Conditioned Cropping} &  & \\
		GAIC~\cite{zeng2020grid}+TransVG~\cite{deng2021transvg} & 0.4230 & 0.5031 \\
		ClipCrop & \bestscore{0.4992} & \bestscore{0.5956} \\
		\bottomrule
	\end{tabular}
\end{table}

\begin{table}[t]
\caption{\textbf{Comparison on the GAIC~\cite{zeng2020grid} dataset.}}
\label{tab:gaic}
	\centering
	\setlength{\tabcolsep}{12pt}
	\begin{tabular}{lcc}
		\toprule
		Method & ACC$_{1/5}$ & ACC$_{1/10}$ \\
		\midrule
		VFN~\cite{chen2017learning} & 26.6 & 40.6 \\
		A2-RL~\cite{li2018a2} & 23.2 & 39.5 \\
		VPN~\cite{wei2018good} & 36.0 & 48.5 \\
		VEN~\cite{wei2018good} & 37.5 & 50.5 \\
		GAIC~\cite{zeng2020grid} & 68.2 & 85.8 \\
		TransView~\cite{pan2021transview} & 69.0 & 85.4 \\
		Jia~\etal~\cite{jia2022rethinking} & 85.0 & 92.6 \\
		\midrule
		ClipCrop-base & \bestscore{87.2} & \bestscore{94.4} \\
		ClipCrop & 79.8 & 89.6 \\
		\bottomrule
	\end{tabular}
\end{table}

\subsection{Results on TextCrop}
\label{sec:results_our_textset}

We followed the test setup of the Horanyi-PR dataset to test on the TextCrop dataset.

The quantitative results are shown in Table~\ref{tab:textcrop}. Unlike Horanyi-PR, each image of TextCrop has two different descriptions. Besides local descriptions, global descriptions exist for most images. Because of this change, the performance of general cropping algorithms becomes better than visual grounding algorithms. ClipCrop still achieves the best performance, demonstrating its ability to provide a crop that accurately reflects the user's intention and has an aesthetic appeal.

We further present the qualitative results on TextCrop in the Fig.~\ref{fig:textcrop_results}. 
Only ClipCrop and the cascaded scheme (GAIC+TransVG) can understand different text queries, such as ``person modeling handbag and dress'' and ``red handbag'' in the first row, to crop images that highlight the corresponding subject. However, ClipCrop is more robust than the cascaded scheme in other cases. With ``groom getting ready'' in the third row and ``math equations'' in the fourth row, ClipCrop again demonstrates its outstanding open vocabulary capability for text-conditioned cropping in the wild. 

\subsection{Results on GAIC}
\label{sec:results_gaic}

\begin{figure}[!t]
  \centering
    \includegraphics[width=.9\linewidth]{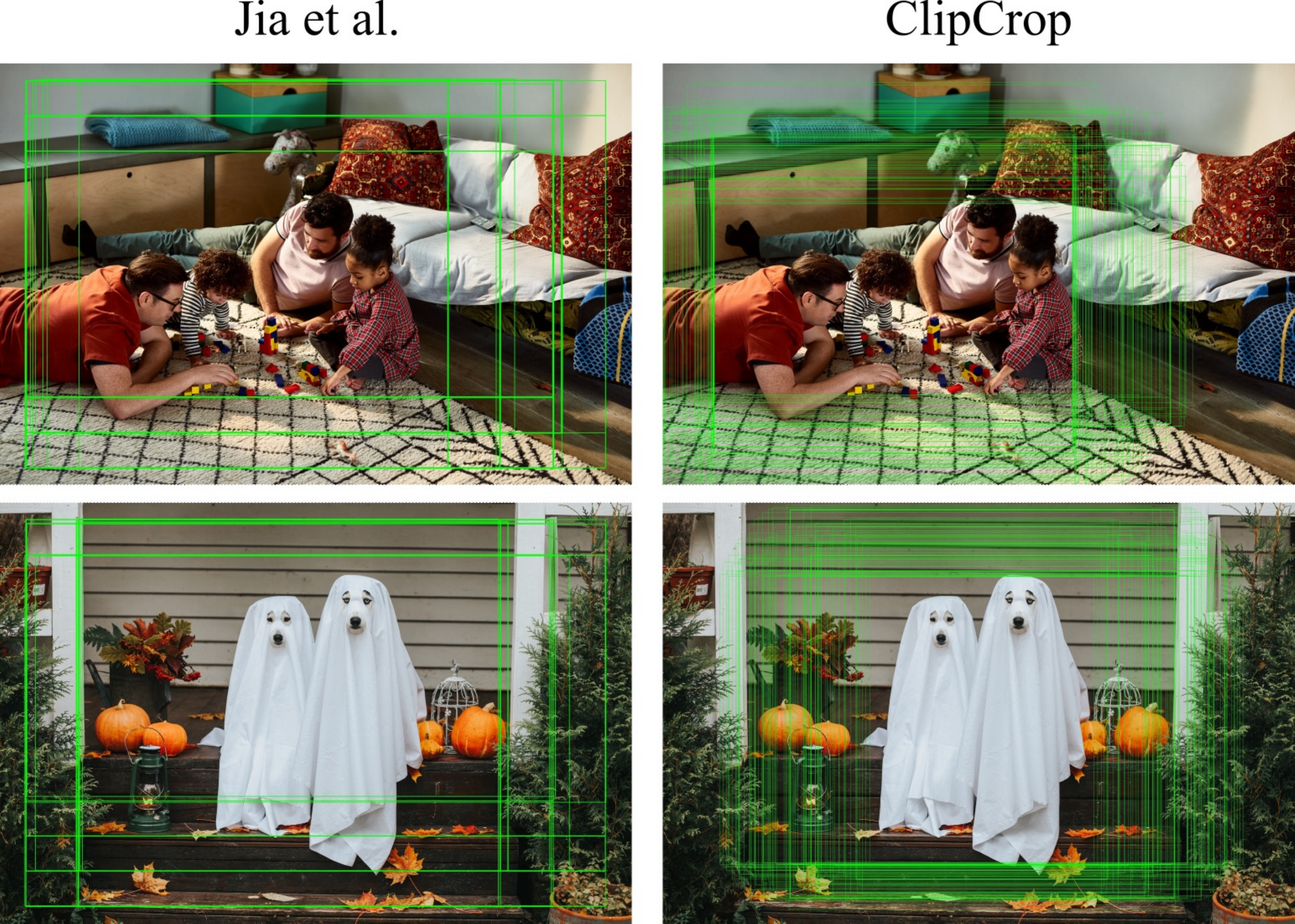}
  \caption{\textbf{Pattern of predicted bounding boxes.}}
  \label{fig:bbox_pattern}
\end{figure}

In addition to the text-conditioned test sets, we also present a comparison with the GAIC~\cite{zeng2020grid}, as shown in Table~\ref{tab:gaic}. Besides ClipCrop, we show a variant called ClipCrop-base, which does not involve the matching of text query but rather regresses directly with query tokens and image tokens. With the boost brought by the introduction of the image embedding from the OWL-ViT encoder, this baseline model achieves state-of-the-art performance without considering text. The reasons why ClipCrop has lower performance under this dataset are twofold. First, the annotated description of GAIC images is subjective, which may introduce some noise to the model. Second, the proposed training strategy results in the generated bounding boxes not overfitting to the grid pattern. As shown in Fig.~\ref{fig:bbox_pattern}, the predicted bounding box pattern of Jia~\etal~\cite{jia2022rethinking} is similar to the ground-truth grid pattern of the GAIC~\cite{zeng2020grid}, while the prediction of ClipCrop is more flexible. Overfitting to the ground-truth pattern is beneficial for the metrics evaluation of this dataset, but harmful for practical applications.

\subsection{One-shot Image-conditioned Querying}
\label{sec:image_query}

\begin{figure}[!t]
  \centering
    \includegraphics[width=\linewidth]{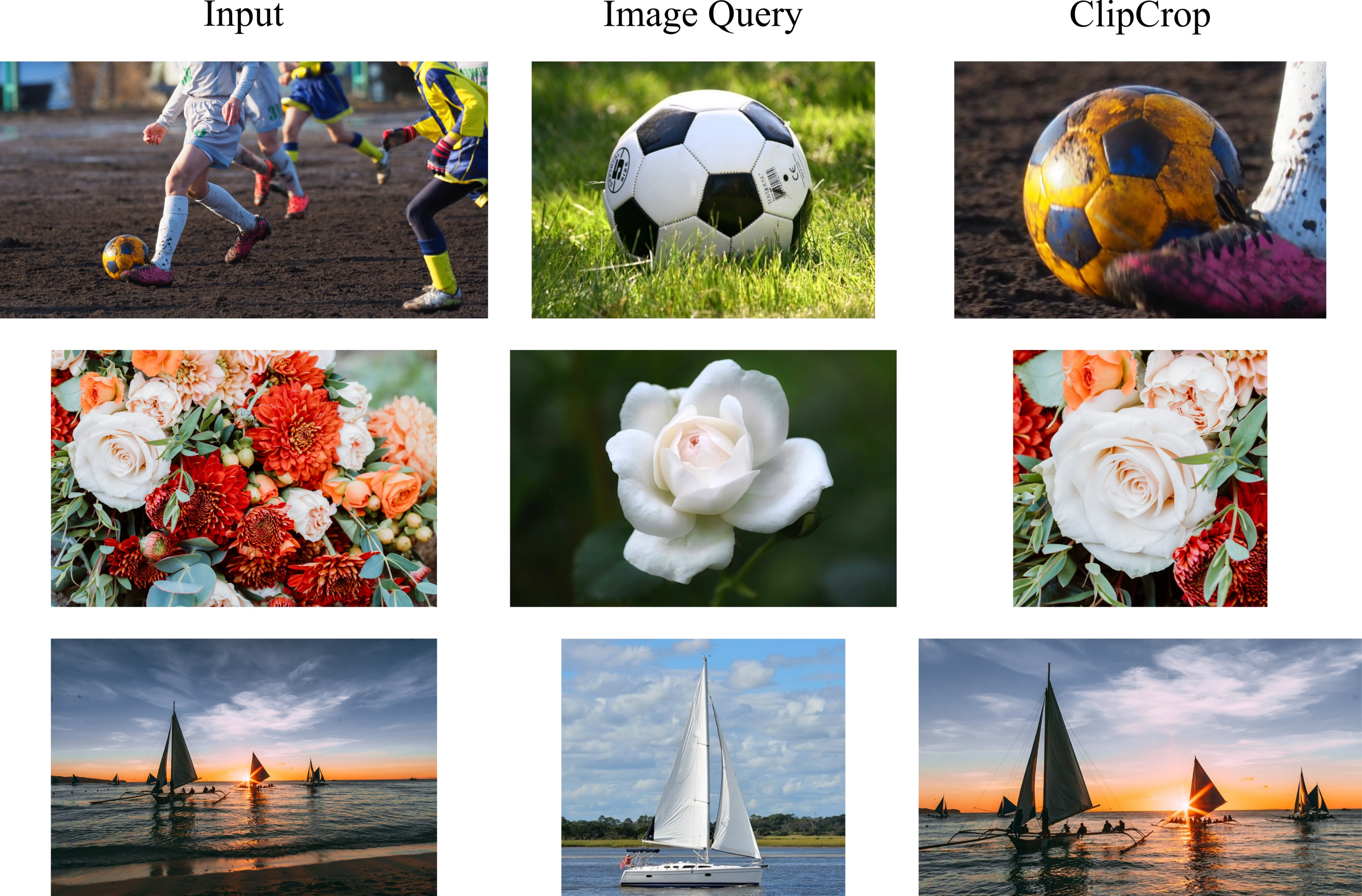}
  \caption{\textbf{Results of one-shot image-conditioned querying.} The size of the crop was adjusted for better viewing.}
  \label{fig:image_query}
\end{figure}

The design of ClipCrop ensures that it inherits the ability from OWL-ViT to perform one-shot image-conditioned querying. We follow the pipeline of OWL-ViT to extract image embeddings from the query image instead of text embeddings to match the initial bounding boxes and image embeddings from the source image. There are some examples for reference in Fig.~\ref{fig:image_query}. The first, second, and third columns represent input image, query image, and crop, respectively. The image-conditioned cropping is a powerful extension to text-conditioned cropping, as it enables cropping the desired areas that are difficult to describe by text but easy to convey by an image.

\begin{table}[t]
\caption{\textbf{Ablation study on Horanyi-PR~\cite{horanyi2022repurposing}.}}
\label{tab:ablation_horanyi-pr}
	\centering
	\setlength{\tabcolsep}{9pt}
	\begin{tabular}{lcc}
		\toprule
		Method & IoU-Mean & IoU-Max \\
		\midrule
		ClipCrop-base & 0.3233 & 0.4742 \\
		ClipCrop-main & 0.4550 & 0.6007 \\
		ClipCrop-key & 0.5170 & 0.6848 \\
		ClipCrop w/o mosaic & 0.5231 & 0.6901 \\
		ClipCrop & 0.5551 & 0.7284 \\
		\bottomrule
	\end{tabular}
\end{table}

\subsection{Ablation Study}
\label{sec:ablation}

To verify the validity of our model design, we conducted an ablation study on Horanyi-PR, as shown in Table~\ref{tab:ablation_horanyi-pr}. ClipCrop w/o mosaic denotes training without mosaic augmentation. ClipCrop-key denotes only using keywords extracted by SpaCy~\cite{Honnibal_spaCy_Industrial-strength_Natural_2020} for matching. Clip-main denotes only using the original text description for matching. ClipCrop-base does not use additional text description as input. The results show that additional text input, combination of original text and keywords, as well as mosaic augmentation all facilitate the text-conditioned cropping. 

\subsection{Limitations}
\label{sec:limitations}

ClipCrop inherits the advantages of VLM while also becoming highly dependent on the pre-trained VLM model. Thus, ClipCrop may include biases introduced by text-image pairs in the contrastive learning stage. Then, because OWL-ViT highly relies on the nouns in the description, the current ClipCrop may not work for descriptions that are too abstract or indirect. 


\section{Conclusion}
\label{sec:conclusion}

In this work, we focus on the challenging problem of how to better reflect the user's intention in the cropping task. Drawing inspiration from vision-language modeling, we presented a novel model, ClipCrop, for text/image-conditioned image cropping. We provided training strategies and designed a network structure that enables ClipCrop to realize conditioned cropping in the wild, with strong open-vocabulary ability. ClipCrop yields superior and robust performance on both the public datasets and the dataset we provide. We believe that introducing pre-trained VLM models into the cropping domain, where data annotation is exceptionally expensive, is a promising direction.


{\small
\bibliographystyle{ieee_fullname}
\bibliography{egbib}

\begin{thebibliography}{10}\itemsep=-1pt

\bibitem{bochkovskiy2020yolov4}
Alexey Bochkovskiy, Chien-Yao Wang, and Hong-Yuan~Mark Liao.
\newblock Yolov4: Optimal speed and accuracy of object detection.
\newblock {\em arXiv preprint arXiv:2004.10934}, 2020.

\bibitem{carion2020end}
Nicolas Carion, Francisco Massa, Gabriel Synnaeve, Nicolas Usunier, Alexander
  Kirillov, and Sergey Zagoruyko.
\newblock End-to-end object detection with transformers.
\newblock In {\em European conference on computer vision}, pages 213--229.
  Springer, 2020.

\bibitem{chen2020uniter}
Yen-Chun Chen, Linjie Li, Licheng Yu, Ahmed El~Kholy, Faisal Ahmed, Zhe Gan, Yu
  Cheng, and Jingjing Liu.
\newblock Uniter: Universal image-text representation learning.
\newblock In {\em European conference on computer vision}, pages 104--120.
  Springer, 2020.

\bibitem{chen2017quantitative}
Yi-Ling Chen, Tzu-Wei Huang, Kai-Han Chang, Yu-Chen Tsai, Hwann-Tzong Chen, and
  Bing-Yu Chen.
\newblock Quantitative analysis of automatic image cropping algorithms: A
  dataset and comparative study.
\newblock In {\em 2017 IEEE winter conference on applications of computer
  vision (WACV)}, pages 226--234. IEEE, 2017.

\bibitem{chen2017learning}
Yi-Ling Chen, Jan Klopp, Min Sun, Shao-Yi Chien, and Kwan-Liu Ma.
\newblock Learning to compose with professional photographs on the web.
\newblock In {\em Proceedings of the 25th ACM international conference on
  Multimedia}, pages 37--45, 2017.

\bibitem{deng2021transvg}
Jiajun Deng, Zhengyuan Yang, Tianlang Chen, Wengang Zhou, and Houqiang Li.
\newblock Transvg: End-to-end visual grounding with transformers.
\newblock In {\em Proceedings of the IEEE/CVF International Conference on
  Computer Vision}, pages 1769--1779, 2021.

\bibitem{devlin2018bert}
Jacob Devlin, Ming-Wei Chang, Kenton Lee, and Kristina Toutanova.
\newblock Bert: Pre-training of deep bidirectional transformers for language
  understanding.
\newblock {\em arXiv preprint arXiv:1810.04805}, 2018.

\bibitem{fang2014automatic}
Chen Fang, Zhe Lin, Radomir Mech, and Xiaohui Shen.
\newblock Automatic image cropping using visual composition, boundary
  simplicity and content preservation models.
\newblock In {\em Proceedings of the 22nd ACM international conference on
  Multimedia}, pages 1105--1108, 2014.

\bibitem{ghiasi2021open}
Golnaz Ghiasi, Xiuye Gu, Yin Cui, and Tsung-Yi Lin.
\newblock Open-vocabulary image segmentation.
\newblock {\em arXiv preprint arXiv:2112.12143}, 2021.

\bibitem{guo2018automatic}
Guanjun Guo, Hanzi Wang, Chunhua Shen, Yan Yan, and Hong-Yuan~Mark Liao.
\newblock Automatic image cropping for visual aesthetic enhancement using deep
  neural networks and cascaded regression.
\newblock {\em IEEE Transactions on Multimedia}, 20(8):2073--2085, 2018.

\bibitem{hong2021composing}
Chaoyi Hong, Shuaiyuan Du, Ke Xian, Hao Lu, Zhiguo Cao, and Weicai Zhong.
\newblock Composing photos like a photographer.
\newblock In {\em Proceedings of the IEEE/CVF Conference on Computer Vision and
  Pattern Recognition}, pages 7057--7066, 2021.

\bibitem{Honnibal_spaCy_Industrial-strength_Natural_2020}
Matthew Honnibal, Ines Montani, Sofie Van~Landeghem, and Adriane Boyd.
\newblock {spaCy: Industrial-strength Natural Language Processing in Python}.
\newblock 2020.

\bibitem{horanyi2022repurposing}
Nora Horanyi, Kedi Xia, Kwang~Moo Yi, Abhishake~Kumar Bojja, Ale{\v{s}}
  Leonardis, and Hyung~Jin Chang.
\newblock Repurposing existing deep networks for caption and aesthetic-guided
  image cropping.
\newblock {\em Pattern Recognition}, 126:108485, 2022.

\bibitem{huang2015automatic}
Jingwei Huang, Huarong Chen, Bin Wang, and Stephen Lin.
\newblock Automatic thumbnail generation based on visual representativeness and
  foreground recognizability.
\newblock In {\em Proceedings of the IEEE International Conference on Computer
  Vision}, pages 253--261, 2015.

\bibitem{jia2021scaling}
Chao Jia, Yinfei Yang, Ye Xia, Yi-Ting Chen, Zarana Parekh, Hieu Pham, Quoc Le,
  Yun-Hsuan Sung, Zhen Li, and Tom Duerig.
\newblock Scaling up visual and vision-language representation learning with
  noisy text supervision.
\newblock In {\em International Conference on Machine Learning}, pages
  4904--4916. PMLR, 2021.

\bibitem{jia2022rethinking}
Gengyun Jia, Huaibo Huang, Chaoyou Fu, and Ran He.
\newblock Rethinking image cropping: Exploring diverse compositions from global
  views.
\newblock In {\em Proceedings of the IEEE/CVF Conference on Computer Vision and
  Pattern Recognition}, pages 2446--2455, 2022.

\bibitem{kamath2021mdetr}
Aishwarya Kamath, Mannat Singh, Yann LeCun, Gabriel Synnaeve, Ishan Misra, and
  Nicolas Carion.
\newblock Mdetr-modulated detection for end-to-end multi-modal understanding.
\newblock In {\em Proceedings of the IEEE/CVF International Conference on
  Computer Vision}, pages 1780--1790, 2021.

\bibitem{kang2014convolutional}
Le Kang, Peng Ye, Yi Li, and David Doermann.
\newblock Convolutional neural networks for no-reference image quality
  assessment.
\newblock In {\em Proceedings of the IEEE conference on computer vision and
  pattern recognition}, pages 1733--1740, 2014.

\bibitem{karpathy2014deep}
Andrej Karpathy, Armand Joulin, and Li~F Fei-Fei.
\newblock Deep fragment embeddings for bidirectional image sentence mapping.
\newblock {\em Advances in neural information processing systems}, 27, 2014.

\bibitem{kim2021vilt}
Wonjae Kim, Bokyung Son, and Ildoo Kim.
\newblock Vilt: Vision-and-language transformer without convolution or region
  supervision.
\newblock In {\em International Conference on Machine Learning}, pages
  5583--5594. PMLR, 2021.

\bibitem{kong2016photo}
Shu Kong, Xiaohui Shen, Zhe Lin, Radomir Mech, and Charless Fowlkes.
\newblock Photo aesthetics ranking network with attributes and content
  adaptation.
\newblock In {\em European conference on computer vision}, pages 662--679.
  Springer, 2016.

\bibitem{krishna2017visual}
Ranjay Krishna, Yuke Zhu, Oliver Groth, Justin Johnson, Kenji Hata, Joshua
  Kravitz, Stephanie Chen, Yannis Kalantidis, Li-Jia Li, David~A Shamma, et~al.
\newblock Visual genome: Connecting language and vision using crowdsourced
  dense image annotations.
\newblock {\em International journal of computer vision}, 123(1):32--73, 2017.

\bibitem{kuhn1955hungarian}
Harold~W Kuhn.
\newblock The hungarian method for the assignment problem.
\newblock {\em Naval research logistics quarterly}, 2(1-2):83--97, 1955.

\bibitem{kuznetsova2020open}
Alina Kuznetsova, Hassan Rom, Neil Alldrin, Jasper Uijlings, Ivan Krasin, Jordi
  Pont-Tuset, Shahab Kamali, Stefan Popov, Matteo Malloci, Alexander
  Kolesnikov, et~al.
\newblock The open images dataset v4.
\newblock {\em International Journal of Computer Vision}, 128(7):1956--1981,
  2020.

\bibitem{lester2021power}
Brian Lester, Rami Al-Rfou, and Noah Constant.
\newblock The power of scale for parameter-efficient prompt tuning.
\newblock {\em arXiv preprint arXiv:2104.08691}, 2021.

\bibitem{li2018a2}
Debang Li, Huikai Wu, Junge Zhang, and Kaiqi Huang.
\newblock A2-rl: Aesthetics aware reinforcement learning for image cropping.
\newblock In {\em Proceedings of the IEEE conference on computer vision and
  pattern recognition}, pages 8193--8201, 2018.

\bibitem{li2020learning}
Debang Li, Junge Zhang, and Kaiqi Huang.
\newblock Learning to learn cropping models for different aspect ratio
  requirements.
\newblock In {\em Proceedings of the IEEE/CVF Conference on Computer Vision and
  Pattern Recognition}, pages 12685--12694, 2020.

\bibitem{li2020composing}
Debang Li, Junge Zhang, Kaiqi Huang, and Ming-Hsuan Yang.
\newblock Composing good shots by exploiting mutual relations.
\newblock In {\em Proceedings of the IEEE/CVF Conference on Computer Vision and
  Pattern Recognition}, pages 4213--4222, 2020.

\bibitem{li2020unicoder}
Gen Li, Nan Duan, Yuejian Fang, Ming Gong, and Daxin Jiang.
\newblock Unicoder-vl: A universal encoder for vision and language by
  cross-modal pre-training.
\newblock In {\em Proceedings of the AAAI Conference on Artificial
  Intelligence}, volume~34, pages 11336--11344, 2020.

\bibitem{li2021align}
Junnan Li, Ramprasaath Selvaraju, Akhilesh Gotmare, Shafiq Joty, Caiming Xiong,
  and Steven Chu~Hong Hoi.
\newblock Align before fuse: Vision and language representation learning with
  momentum distillation.
\newblock {\em Advances in neural information processing systems},
  34:9694--9705, 2021.

\bibitem{li2022grounded}
Liunian~Harold Li, Pengchuan Zhang, Haotian Zhang, Jianwei Yang, Chunyuan Li,
  Yiwu Zhong, Lijuan Wang, Lu Yuan, Lei Zhang, Jenq-Neng Hwang, et~al.
\newblock Grounded language-image pre-training.
\newblock In {\em Proceedings of the IEEE/CVF Conference on Computer Vision and
  Pattern Recognition}, pages 10965--10975, 2022.

\bibitem{li2020oscar}
Xiujun Li, Xi Yin, Chunyuan Li, Pengchuan Zhang, Xiaowei Hu, Lei Zhang, Lijuan
  Wang, Houdong Hu, Li Dong, Furu Wei, et~al.
\newblock Oscar: Object-semantics aligned pre-training for vision-language
  tasks.
\newblock In {\em European Conference on Computer Vision}, pages 121--137.
  Springer, 2020.

\bibitem{liu2021pre}
Pengfei Liu, Weizhe Yuan, Jinlan Fu, Zhengbao Jiang, Hiroaki Hayashi, and
  Graham Neubig.
\newblock Pre-train, prompt, and predict: A systematic survey of prompting
  methods in natural language processing.
\newblock {\em arXiv preprint arXiv:2107.13586}, 2021.

\bibitem{liu2016ssd}
Wei Liu, Dragomir Anguelov, Dumitru Erhan, Christian Szegedy, Scott Reed,
  Cheng-Yang Fu, and Alexander~C Berg.
\newblock Ssd: Single shot multibox detector.
\newblock In {\em European conference on computer vision}, pages 21--37.
  Springer, 2016.

\bibitem{loshchilov2017decoupled}
Ilya Loshchilov and Frank Hutter.
\newblock Decoupled weight decay regularization.
\newblock {\em arXiv preprint arXiv:1711.05101}, 2017.

\bibitem{lu2019vilbert}
Jiasen Lu, Dhruv Batra, Devi Parikh, and Stefan Lee.
\newblock Vilbert: Pretraining task-agnostic visiolinguistic representations
  for vision-and-language tasks.
\newblock {\em Advances in neural information processing systems}, 32, 2019.

\bibitem{lu202012}
Jiasen Lu, Vedanuj Goswami, Marcus Rohrbach, Devi Parikh, and Stefan Lee.
\newblock 12-in-1: Multi-task vision and language representation learning.
\newblock In {\em Proceedings of the IEEE/CVF Conference on Computer Vision and
  Pattern Recognition}, pages 10437--10446, 2020.

\bibitem{lu2020weakly}
Peng Lu, Jiahui Liu, Xujun Peng, and Xiaojie Wang.
\newblock Weakly supervised real-time image cropping based on aesthetic
  distributions.
\newblock In {\em Proceedings of the 28th ACM International Conference on
  Multimedia}, pages 120--128, 2020.

\bibitem{lu2014rapid}
Xin Lu, Zhe Lin, Hailin Jin, Jianchao Yang, and James~Z Wang.
\newblock Rapid: Rating pictorial aesthetics using deep learning.
\newblock In {\em Proceedings of the 22nd ACM international conference on
  Multimedia}, pages 457--466, 2014.

\bibitem{mao2016generation}
Junhua Mao, Jonathan Huang, Alexander Toshev, Oana Camburu, Alan~L Yuille, and
  Kevin Murphy.
\newblock Generation and comprehension of unambiguous object descriptions.
\newblock In {\em Proceedings of the IEEE conference on computer vision and
  pattern recognition}, pages 11--20, 2016.

\bibitem{minderer2022simple}
Matthias Minderer, Alexey Gritsenko, Austin Stone, Maxim Neumann, Dirk
  Weissenborn, Alexey Dosovitskiy, Aravindh Mahendran, Anurag Arnab, Mostafa
  Dehghani, Zhuoran Shen, et~al.
\newblock Simple open-vocabulary object detection with vision transformers.
\newblock {\em arXiv preprint arXiv:2205.06230}, 2022.

\bibitem{mokady2021clipcap}
Ron Mokady, Amir Hertz, and Amit~H Bermano.
\newblock Clipcap: Clip prefix for image captioning.
\newblock {\em arXiv preprint arXiv:2111.09734}, 2021.

\bibitem{ni2013learning}
Bingbing Ni, Mengdi Xu, Bin Cheng, Meng Wang, Shuicheng Yan, and Qi Tian.
\newblock Learning to photograph: A compositional perspective.
\newblock {\em IEEE Transactions on Multimedia}, 15(5):1138--1151, 2013.

\bibitem{nishiyama2009sensation}
Masashi Nishiyama, Takahiro Okabe, Yoichi Sato, and Imari Sato.
\newblock Sensation-based photo cropping.
\newblock In {\em Proceedings of the 17th ACM international conference on
  Multimedia}, pages 669--672, 2009.

\bibitem{pan2021transview}
Zhiyu Pan, Zhiguo Cao, Kewei Wang, Hao Lu, and Weicai Zhong.
\newblock Transview: Inside, outside, and across the cropping view boundaries.
\newblock In {\em Proceedings of the IEEE/CVF International Conference on
  Computer Vision}, pages 4218--4227, 2021.

\bibitem{paszke2019pytorch}
Adam Paszke, Sam Gross, Francisco Massa, Adam Lerer, James Bradbury, Gregory
  Chanan, Trevor Killeen, Zeming Lin, Natalia Gimelshein, Luca Antiga, et~al.
\newblock Pytorch: An imperative style, high-performance deep learning library.
\newblock {\em arXiv preprint arXiv:1912.01703}, 2019.

\bibitem{pham2021combined}
Hieu Pham, Zihang Dai, Golnaz Ghiasi, Hanxiao Liu, Adams~Wei Yu, Minh-Thang
  Luong, Mingxing Tan, and Quoc~V Le.
\newblock Combined scaling for zero-shot transfer learning.
\newblock {\em arXiv preprint arXiv:2111.10050}, 2021.

\bibitem{plummer2015flickr30k}
Bryan~A Plummer, Liwei Wang, Chris~M Cervantes, Juan~C Caicedo, Julia
  Hockenmaier, and Svetlana Lazebnik.
\newblock Flickr30k entities: Collecting region-to-phrase correspondences for
  richer image-to-sentence models.
\newblock In {\em Proceedings of the IEEE international conference on computer
  vision}, pages 2641--2649, 2015.

\bibitem{radford2021learning}
Alec Radford, Jong~Wook Kim, Chris Hallacy, Aditya Ramesh, Gabriel Goh,
  Sandhini Agarwal, Girish Sastry, Amanda Askell, Pamela Mishkin, Jack Clark,
  et~al.
\newblock Learning transferable visual models from natural language
  supervision.
\newblock In {\em International Conference on Machine Learning}, pages
  8748--8763. PMLR, 2021.

\bibitem{rao2022denseclip}
Yongming Rao, Wenliang Zhao, Guangyi Chen, Yansong Tang, Zheng Zhu, Guan Huang,
  Jie Zhou, and Jiwen Lu.
\newblock Denseclip: Language-guided dense prediction with context-aware
  prompting.
\newblock In {\em Proceedings of the IEEE/CVF Conference on Computer Vision and
  Pattern Recognition}, pages 18082--18091, 2022.

\bibitem{redmon2018yolov3}
Joseph Redmon and Ali Farhadi.
\newblock Yolov3: An incremental improvement.
\newblock {\em arXiv preprint arXiv:1804.02767}, 2018.

\bibitem{sanghi2022clip}
Aditya Sanghi, Hang Chu, Joseph~G Lambourne, Ye Wang, Chin-Yi Cheng, Marco
  Fumero, and Kamal~Rahimi Malekshan.
\newblock Clip-forge: Towards zero-shot text-to-shape generation.
\newblock In {\em Proceedings of the IEEE/CVF Conference on Computer Vision and
  Pattern Recognition}, pages 18603--18613, 2022.

\bibitem{shao2019objects365}
Shuai Shao, Zeming Li, Tianyuan Zhang, Chao Peng, Gang Yu, Xiangyu Zhang, Jing
  Li, and Jian Sun.
\newblock Objects365: A large-scale, high-quality dataset for object detection.
\newblock In {\em Proceedings of the IEEE/CVF international conference on
  computer vision}, pages 8430--8439, 2019.

\bibitem{shi2022dynamic}
Fengyuan Shi, Ruopeng Gao, Weilin Huang, and Limin Wang.
\newblock Dynamic mdetr: A dynamic multimodal transformer decoder for visual
  grounding.
\newblock {\em arXiv preprint arXiv:2209.13959}, 2022.

\bibitem{su2019vl}
Weijie Su, Xizhou Zhu, Yue Cao, Bin Li, Lewei Lu, Furu Wei, and Jifeng Dai.
\newblock Vl-bert: Pre-training of generic visual-linguistic representations.
\newblock {\em arXiv preprint arXiv:1908.08530}, 2019.

\bibitem{wada2018labelme}
Kentaro Wada.
\newblock labelme: Image polygonal annotation with python.
\newblock \url{https://github.com/wkentaro/labelme}, 2018.

\bibitem{wang2021simvlm}
Zirui Wang, Jiahui Yu, Adams~Wei Yu, Zihang Dai, Yulia Tsvetkov, and Yuan Cao.
\newblock Simvlm: Simple visual language model pretraining with weak
  supervision.
\newblock {\em arXiv preprint arXiv:2108.10904}, 2021.

\bibitem{wei2018good}
Zijun Wei, Jianming Zhang, Xiaohui Shen, Zhe Lin, Radomir Mech, Minh Hoai, and
  Dimitris Samaras.
\newblock Good view hunting: Learning photo composition from dense view pairs.
\newblock In {\em Proceedings of the IEEE conference on computer vision and
  pattern recognition}, pages 5437--5446, 2018.

\bibitem{wu2022wav2clip}
Ho-Hsiang Wu, Prem Seetharaman, Kundan Kumar, and Juan~Pablo Bello.
\newblock Wav2clip: Learning robust audio representations from clip.
\newblock In {\em ICASSP 2022-2022 IEEE International Conference on Acoustics,
  Speech and Signal Processing (ICASSP)}, pages 4563--4567. IEEE, 2022.

\bibitem{xu2021videoclip}
Hu Xu, Gargi Ghosh, Po-Yao Huang, Dmytro Okhonko, Armen Aghajanyan, Florian
  Metze, Luke Zettlemoyer, and Christoph Feichtenhofer.
\newblock Videoclip: Contrastive pre-training for zero-shot video-text
  understanding.
\newblock {\em arXiv preprint arXiv:2109.14084}, 2021.

\bibitem{xu2022groupvit}
Jiarui Xu, Shalini De~Mello, Sifei Liu, Wonmin Byeon, Thomas Breuel, Jan Kautz,
  and Xiaolong Wang.
\newblock Groupvit: Semantic segmentation emerges from text supervision.
\newblock In {\em Proceedings of the IEEE/CVF Conference on Computer Vision and
  Pattern Recognition}, pages 18134--18144, 2022.

\bibitem{xu2015show}
Kelvin Xu, Jimmy Ba, Ryan Kiros, Kyunghyun Cho, Aaron Courville, Ruslan
  Salakhudinov, Rich Zemel, and Yoshua Bengio.
\newblock Show, attend and tell: Neural image caption generation with visual
  attention.
\newblock In {\em International conference on machine learning}, pages
  2048--2057. PMLR, 2015.

\bibitem{yan2013learning}
Jianzhou Yan, Stephen Lin, Sing Bing~Kang, and Xiaoou Tang.
\newblock Learning the change for automatic image cropping.
\newblock In {\em Proceedings of the IEEE conference on computer vision and
  pattern recognition}, pages 971--978, 2013.

\bibitem{yang2022unified}
Jianwei Yang, Chunyuan Li, Pengchuan Zhang, Bin Xiao, Ce Liu, Lu Yuan, and
  Jianfeng Gao.
\newblock Unified contrastive learning in image-text-label space.
\newblock In {\em Proceedings of the IEEE/CVF Conference on Computer Vision and
  Pattern Recognition}, pages 19163--19173, 2022.

\bibitem{yang2020improving}
Zhengyuan Yang, Tianlang Chen, Liwei Wang, and Jiebo Luo.
\newblock Improving one-stage visual grounding by recursive sub-query
  construction.
\newblock In {\em European Conference on Computer Vision}, pages 387--404.
  Springer, 2020.

\bibitem{yu2018mattnet}
Licheng Yu, Zhe Lin, Xiaohui Shen, Jimei Yang, Xin Lu, Mohit Bansal, and
  Tamara~L Berg.
\newblock Mattnet: Modular attention network for referring expression
  comprehension.
\newblock In {\em Proceedings of the IEEE Conference on Computer Vision and
  Pattern Recognition}, pages 1307--1315, 2018.

\bibitem{yu2016modeling}
Licheng Yu, Patrick Poirson, Shan Yang, Alexander~C Berg, and Tamara~L Berg.
\newblock Modeling context in referring expressions.
\newblock In {\em European Conference on Computer Vision}, pages 69--85.
  Springer, 2016.

\bibitem{zeng2019reliable}
Hui Zeng, Lida Li, Zisheng Cao, and Lei Zhang.
\newblock Reliable and efficient image cropping: A grid anchor based approach.
\newblock In {\em Proceedings of the IEEE/CVF conference on computer vision and
  pattern recognition}, pages 5949--5957, 2019.

\bibitem{zeng2020grid}
Hui Zeng, Lida Li, Zisheng Cao, and Lei Zhang.
\newblock Grid anchor based image cropping: A new benchmark and an efficient
  model.
\newblock {\em IEEE Transactions on Pattern Analysis and Machine Intelligence},
  2020.

\bibitem{zhai2022lit}
Xiaohua Zhai, Xiao Wang, Basil Mustafa, Andreas Steiner, Daniel Keysers,
  Alexander Kolesnikov, and Lucas Beyer.
\newblock Lit: Zero-shot transfer with locked-image text tuning.
\newblock In {\em Proceedings of the IEEE/CVF Conference on Computer Vision and
  Pattern Recognition}, pages 18123--18133, 2022.

\bibitem{zhang2022glipv2}
Haotian Zhang, Pengchuan Zhang, Xiaowei Hu, Yen-Chun Chen, Liunian~Harold Li,
  Xiyang Dai, Lijuan Wang, Lu Yuan, Jenq-Neng Hwang, and Jianfeng Gao.
\newblock Glipv2: Unifying localization and vision-language understanding.
\newblock {\em arXiv preprint arXiv:2206.05836}, 2022.

\bibitem{zhong2022regionclip}
Yiwu Zhong, Jianwei Yang, Pengchuan Zhang, Chunyuan Li, Noel Codella,
  Liunian~Harold Li, Luowei Zhou, Xiyang Dai, Lu Yuan, Yin Li, et~al.
\newblock Regionclip: Region-based language-image pretraining.
\newblock In {\em Proceedings of the IEEE/CVF Conference on Computer Vision and
  Pattern Recognition}, pages 16793--16803, 2022.

\end{thebibliography}
}

\clearpage
\appendix
\section{TextCrop}
\begin{figure*}[!t]
\centering
    \includegraphics[width=\linewidth]{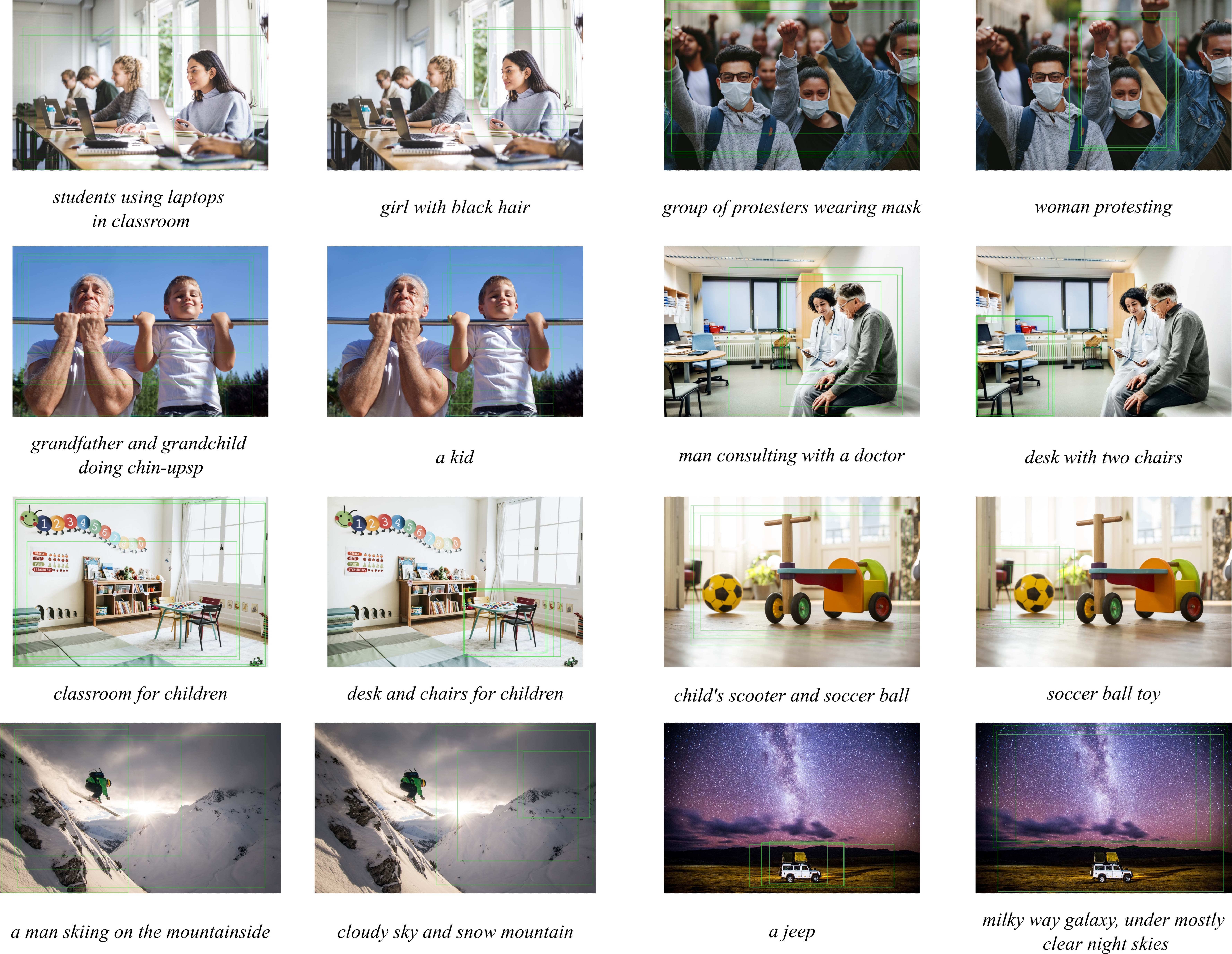}
    \caption{\textbf{Samples from TextCrop.}}
    \label{fig:textcrop_samples}
\end{figure*}
There are 208 images in the TextCrop evaluation set. We show some examples in Fig.~\ref{fig:textcrop_samples} for reference. Text labels and bounding boxes labels are annotated by different annotators. Each image has two distinct text labels, and each text label has four bounding boxes (green boxes in Fig.~\ref{fig:textcrop_samples}) from different annotators. The bounding boxes are labeled by using the open source software labelme~\cite{wada2018labelme}.

\section{Training details}
We optimize the loss using AdamW~\cite{loshchilov2017decoupled} in the PyTorch framework~\cite{paszke2019pytorch}. The learning rate is scheduled by the cosine scheduler from $1\times10^{-4}$ to $1\times10^{-6}$. In addition to mosaics, common data augmentation operations, including color jittering and flipping, are used. We use the encoders from the ViT-B/32 version of OWL-ViT~\cite{minderer2022simple}. The decoder includes 6 attention layers. The dimension of each layer is 512 and the number of heads is 8. The hidden dimension of the MLP layers is 2048. The number of query tokens is 90. We follow Jia~\etal~\cite{jia2022rethinking} to set loss functions. We replace the loss function of quality score with Smooth $\mathcal{L}_{1}$ loss and directly map the quality score to $0\sim 1$ without truncation. We train ClipCrop on GAIC~\cite{zeng2020grid} with a batch size of 32 for 180 epochs, on 8 NVIDIA Tesla V100 GPUs.


\section{Experiments with image captioning labels}
When we train ClipCrop, we provided a text label from human annotators for each image. A valuable question is what happens if the text generated by the image captioning algorithm is used as a text label for training the model? To answer this question, we used ClipCap~\cite{mokady2021clipcap} to generate text labels instead of the real labels of human annotators to train ClipCrop. The results are denoted as ``ClipCrop w/o real'' in Table~\ref{tab:ablation_real}. We can see real text labels do help improve performance, while ClipCrop using generated text labels can still outperform the prior art.

\begin{table*}[!h]
\caption{\textbf{Ablation studies of real text annotation.}}
\label{tab:ablation_real}
	\centering
	\setlength{\tabcolsep}{12pt}
	\begin{tabular}{lcccc}
		\toprule
		& \multicolumn{2}{c}{Horanyi-PR~\cite{horanyi2022repurposing}} & \multicolumn{2}{c}{TextCrop} \\
		\cmidrule(r){2-3}
		\cmidrule(r){4-5}
		Method & IoU-Mean & IoU-Max & IoU-Mean & IoU-Max\\
		\midrule
		ClipCrop w/o real & 0.5094 & 0.6753 & 0.4836 & 0.5801 \\
		ClipCrop & 0.5551 & 0.7284 & 0.4992 & 0.5956 \\
		\bottomrule
	\end{tabular}
\end{table*}

\end{document}